\documentclass[letterpaper, 10 pt, conference]{format/ieeeconf}  

\IEEEoverridecommandlockouts                              

\let\labelindent\relax
\usepackage{enumitem}

\usepackage{graphics} 
\usepackage{graphicx} 
\usepackage{amsmath} 
\usepackage{amssymb}  
\usepackage{color}
\usepackage{hyperref}
\usepackage{lipsum}
\usepackage{multirow}
\usepackage[singlelinecheck=false,justification=justified]{caption}
\usepackage{subcaption}
\usepackage[noadjust,compress]{cite}
\usepackage[ruled,vlined,linesnumbered,noend]{algorithm2e}
\usepackage{etoolbox}
\usepackage{xspace}
\usepackage[table]{xcolor}
\usepackage{comment}
\usepackage{soul}
\usepackage{adjustbox}

\def\bl#1{\textcolor{blue}{#1}}

\newcommand{\name}{{\tt PROBE}\xspace}

\font\titlefont=ptmb at 15.8pt
\title{\titlefont
{\tt PROBE}: Proprioceptive Obstacle Detection and Estimation\\ while Navigating in Clutter
}

\author{Dhruv Metha Ramesh$^{1}$, Aravind Sivaramakrishnan$^{2}$, Shreesh Keskar$^{1}$, \\Kostas E. Bekris$^{3}$, Jingjin Yu$^{1}$, Abdeslam Boularias$^{1}$
\thanks{$^{1}$Dept. of Computer Science, Rutgers University, NJ, USA. $^{2}$A.S. is affiliated with Amazon.com Inc. but the work related to this paper was performed while at Rutgers University. $^{3}$K. E. Bekris holds concurrent appointments as a Professor at Rutgers University and as an Amazon Scholar. This paper describes work performed at Rutgers and is not associated with Amazon. This work is partly
supported by NSF awards 1846043 and 2132972. Corresponding author e-mail: {\tt dhruv.metha@rutgers.edu}.}%
}

\begin{document}

\maketitle
\thispagestyle{empty}
\pagestyle{empty}

\begin{abstract}
In critical applications, including search-and-rescue in degraded environments, blockages can be prevalent and prevent the effective deployment of certain sensing modalities, particularly vision, due to occlusion and the constrained range of view of onboard camera sensors. 
To enable robots to tackle these challenges, we propose a new approach, \bl{Pr}oprioceptive \bl{Ob}stacle Detection and \bl{E}stimation while navigating in clutter (\name), which instead relies only on the robot's proprioception to infer the presence or absence of occluded rectangular obstacles while predicting their dimensions and poses in {\tt SE(2)}. The proposed approach is a Transformer neural network that receives as input a history of applied torques and sensed whole-body movements of the robot and returns a parameterized representation of the obstacles in the environment. The effectiveness of \name is evaluated on simulated environments in Isaac Gym and with a real Unitree Go1 quadruped robot. The project webpage can be found at \href{https://dhruvmetha.github.io/legged-probe/}{\textcolor{blue}{https://dhruvmetha.github.io/legged-probe/}}.
\end{abstract}



\section{Introduction}
\label{sec:introduction}

In a dark environment, humans can effectively navigate by relying on mechanical interactions with the fixed and movable obstacles present. While navigating in such an environment, humans construct in real-time a mental representation of these objects, including their dimensions and poses, entirely through \emph{contact force-based interactive sensing}.
Similarly, modeling the movability of objects on the fly is an important skill for autonomous robots deployed in open environments, such as those encountered in search-and-rescue (SAR) operations, unmanned exploration missions, and debris removal~\cite{BoulariasBS14,Haynes:2017, Atkeson2018, johnson2015team, luo2014robust, pratt2013darpa,RSS2020Changkyu}. These environments can contain objects with diverse physical properties in terms of shape and materials. Therefore, some level of on-demand object property identification can help guide navigation.

Inferring the shapes of objects in cluttered scenes is extremely challenging because of occlusions, which only permit a partial view of the surrounding environment. For instance, in a degraded environment, objects can be hidden behind rubble and thus cannot be observed directly. Visual input may also be extremely limited or impossible in certain critical situations, such as navigating through dense smoke. 

\begin{figure}
    \centering
    \includegraphics[width=0.48\textwidth]{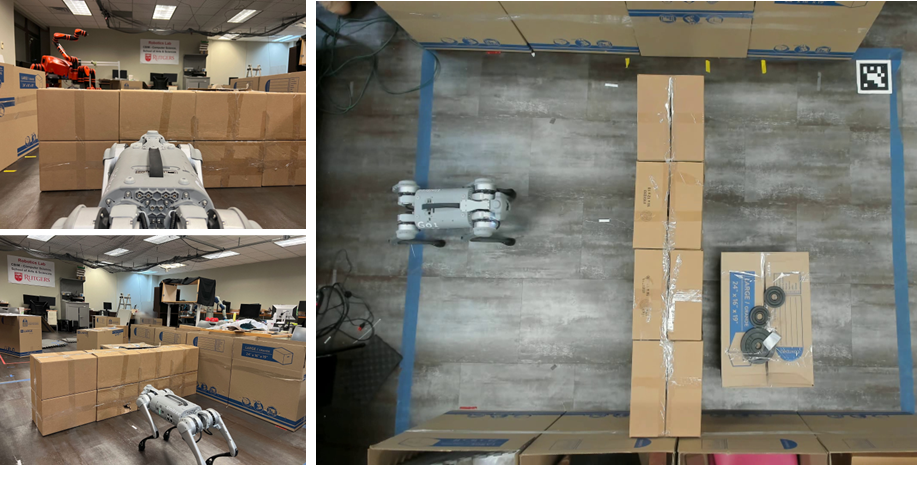}
    \vspace{-.25in}
    \caption{\small The setup considered in \name involves a Go1 robot dog and obstacles that can potentially obstruct its path. Some planar obstacles, such as the long frontal box in the image, are movable, while others are fixed to the ground. A transformer network reconstructs the locations and sizes of the obstacles, including the occluded ones, from a history of proprioceptive data that the robot receives while exploring the scene without vision.  }
    \label{fig:intro-figure}
    \vspace{-0.25in}
\end{figure}

This work addresses the problem of partial scene reconstruction for a legged robot using only proprioception data. We consider a setup (Fig~\ref{fig:intro-figure}) where a legged robot is deployed in an unknown environment and tasked to navigate to a target location without access to visual inputs. 
A learned high-level navigation policy generates desired velocities for the robot, unaware of any obstacles present in the environment, and a learned low-level locomotion controller translates them into joint positions.
A Transformer-based Obstacle Reconstruction Module (ORM) receives as inputs the history of the robot's {\tt SE(2)} pose, the desired velocity commands, its joint positions, velocities, and torques at each time-step. It returns a reconstructed environment representation at each time-step, which is gradually fine-tuned as the robot navigates in its environment. Using only this proprioceptive feedback, the ORM can predict the existence of planar rectangular obstacles, their 2D dimensions and positions. In addition, the ORM can detect whether the encountered obstacles are fixed or movable, and in the latter case, further detect other obstacles hidden behind, and what their properties are. The capability to sense objects hidden behind movable large objects cannot be attained through vision. 
This information can enable the robot to use a high-level reasoning framework to navigate among movable objects \cite{Stilman-2007-9848} by clearing the path unlocking regions to explore.

In summary, this work brings forth the following main contributions:
\begin{itemize}[leftmargin=4mm]
\item The construction of a neural architecture, \name,
capable of rapidly predicting an actionable representation of the robot's environment using only the history of a robot's proprioceptive states during navigation in that environment, without any vision sensors, and
\item A comprehensive empirical study of the proposed system in simulation and on a real quadruped, demonstrating the effectiveness of \name.
\end{itemize}

\section{Related Work}
\label{sec:prelims}


\textbf{Object/Scene Reconstruction} has been extensively studied using vision-based methods that rely on RGB-D images for volumetric shape completion of partially occluded objects~\cite{Varley2016ShapeCE,changkyu-RAL2019,yan2022shapeformer}.
To infer 3D shapes in clutter without vision, prior physics-based methods like~\cite{changkyu-RAL2019} perform computationally expensive searches over large shape spaces based on their observed interactions. 
The proposed ORM can rapidly predict planar obstacle properties directly from proprioceptive data, circumventing intensive computations. 
Other related work focusing on scene reconstruction from force sensing~\cite{9718152} do not deal with {\it nested} object interactions considered in this paper's problem setup.

{\bf Contact Sensing} is crucial for scene reconstruction from force measurements. Early attempts solve the force-moment balance and surface equations while assuming a soft finger and convex shape of the contacted object~\cite{Bicchi1990ContactSF}. These assumptions were relaxed in subsequent works~\cite{DBLP:conf/iros/KuritaST12,DBLP:journals/arobots/LiuNPBBGSA15}. Some works addressed the problem of locating contact points on manipulators~\cite{DeLuca2006CollisionDA}, which is closer to the presented work on legged robots.

{\bf Mapping and Navigation} involves building a map of a mobile robot's unknown environment while the robot is navigating an environment. The simultaneous localization and mapping (SLAM) problem is extensively studied in mobile robotics, with most solutions requiring visual inputs in the form of RGB or LIDAR images~\cite{durrant2006simultaneous,stachniss2016simultaneous,fuentes2015visual}. In this paper, the robot navigates under uncertainty, only observing the environment through contact, a variant of the {\it Blindfolded Traveler’s Problem}~\cite{Saund2019TheBR}. Proposed solutions assume rigid and fixed obstacles~\cite{9384158,9392269,8794262}, which is different from our setup, which contains both fixed and movable obstacles.


{\bf Legged Robot Navigation} utilizes proprioceptive feedback to adapt locomotion control on challenging terrains in real-time \cite{https://doi.org/10.48550/arxiv.2107.04034, https://doi.org/10.48550/arxiv.2211.07638, https://doi.org/10.48550/arxiv.2205.02824}. Unlike methods that implicitly encode environmental conditions, the proposed approach leverages proprioception to reconstruct the environment by inferring the presence and properties of obstacles. This shifts the use of proprioception from internal adaptation to external environment mapping, addressing a different aspect of autonomous navigation in complex environments.



\section{Problem and Environment Setup}
\begin{figure*}[ht!]
    \centering \includegraphics[width=\textwidth]{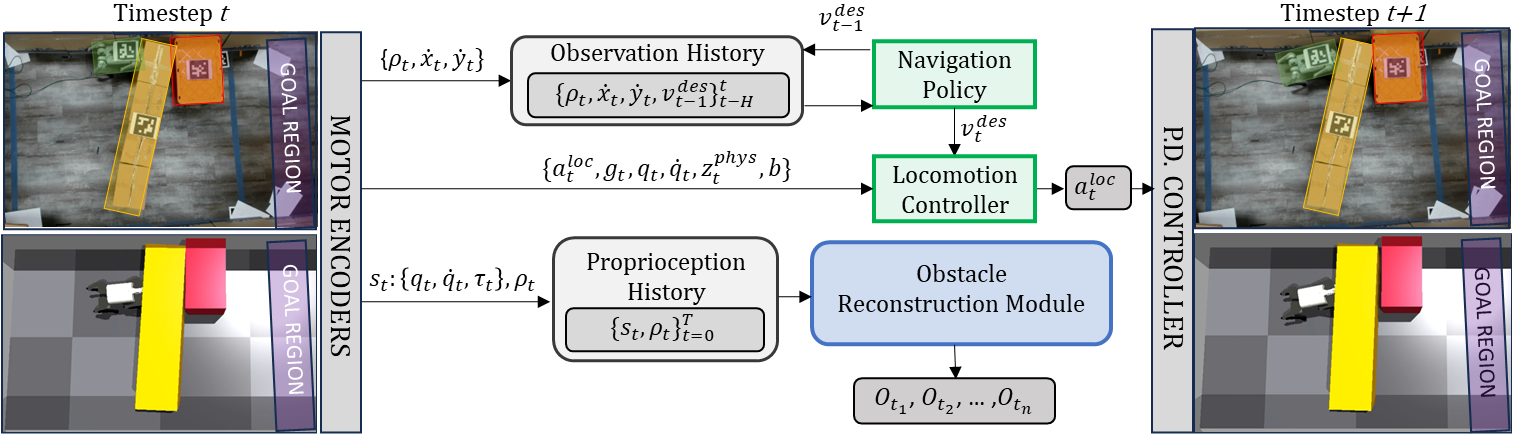}
    \caption{(Left) Environment setup in reality (top) and simulation (bottom), best viewed in color. The robot's workspace is bounded in a box of dimensions $w_\text{env} \times l_\text{env}$, and the goal region (purple) is defined as the set of all locations that satisfy $\{x > K \}$, i.e., all locations that are beyond $K$ meters in the direction of the robot's initial orientation. The yellow long obstacle is movable, while the red obstacle is static. The robot has to move the yellow box in front of it to reach the goal region. (Middle-Right) A hierarchical control policy is executed to navigate the robot so it may explore the environment's properties and successfully reach the goal region. Concurrently, the proposed obstacle reconstruction module (ORM) is a Transformer neural network that uses localization and proprioception history from the robot to predict object position and dimensions through interaction. }
    \label{fig:inference-pipeline}
    \vspace{-0.2in}
\end{figure*}

We address the problem of Obstacle Prediction while navigating in clutter. A legged robot negotiates a 2D workspace to reach a goal region (Fig.~\ref{fig:inference-pipeline}) without visual inputs.  The environment contains varying numbers of box-shaped obstacles that are either mobile (e.g., movable by pushing) or immobile. The locations and sizes of the obstacles in the environment are unknown \emph{a priori}. While navigating, the robot manipulates or comes into contact with a subset of the obstacles in its workspace. Given the robot's proprioceptive state history, the environment is partially reconstructed by predicting the locations, sizes, and mobility of the obstacles it has encountered, either directly or indirectly, through nested interactions. Nested interactions refer to cases when the robot pushes a moving obstacle, which in turn pushes a second obstacle that is behind it.

\noindent\textbf{Robotic Platform.} This work uses the Unitree Go1-EDU legged robot. Locomotion is performed by controlling its twelve joints. The proprioception signal at time-step $t$ is  $s_t = \{(q_i, \dot{q}_i, \tau_i)\}_{i=1}^{12}$, where $(q_i, \dot{q}_i)$ are the position and velocity of joint $i$, and $\tau_i$ is the torque applied on joint $i$. The pose and velocity of the robot at time $t$ is $\rho_t = (x_t, y_t, \theta_t)$ and $\dot{\rho}_t = (\dot{x}_t, \dot{y}_t, \dot{\theta_t})$, where $(x_t,y_t,\theta_t) \in {\tt SE}(2)$ is the 2D translational and rotational transformations w.r.t. the robot's initial position $A$, and $(\dot{x}_t, \dot{y}_t, \dot{\theta}_t)$ are the linear and angular velocity components. 

Both movable and static obstacles are placed randomly in the workspace and have unknown friction, mass, and size. \name's scene estimation module represents each obstacle as $\mathcal{O}_i = (\mathbb{I}^\text{static}_i, x_i, y_i, \theta_i, w_i, l_i)$. The binary variable $\mathbb{I}^\text{static}_i$ indicates whether the obstacle is static ($1)$ or movable ($0$). $(x_i,y_i,\theta_i) \in {\tt SE}(2)$ defines the pose of the obstacle relative to the robot's initial position $A$ (typically the origin), and $(w_i,l_i)$ defines the box's width and length respectively. The set of all obstacles in the environment with $N$ obstacles is $\mathcal{O} = \{\mathcal{O}_i\}_{i=1}^N$.

\section{Proposed Method}
\label{sec:proposed}


\subsection{Low-level Locomotion Controller}
The locomotion controller $\pi_\text{loc}$ \cite{margolis2022walk} is a neural network trained in an obstacle-free environment using Proximal Policy Optimization (PPO) \cite{schulman2017proximal}. $\pi_\text{loc}$ tracks velocity and gait commands, enabling the robot to walk at desired speeds while maintaining a commanded gait. The velocity commands $v^\text{des}$ are specified within the robot's $\tt SE(2)$ frame. Gait commands $b$ include the robot's stepping frequency, body height, and stance. The input observations $o^\text{loc} $ to the policy are the gravity vector $g$ in the robot's frame, the robot's proprioceptive states -- joint positions $q$, joint velocities $\dot{q}$, the previous action $a^\text{loc}_\text{prev}$, and a latent physics parameter $z^\text{phys}$. The commands $c={(v^\text{des}, b)}$ are also provided as input. The output action $a^\text{loc}$ is the target joint position for each of the robot's joints. These target joint positions are converted to joint torques $\tau$ using the robot's built-in PD controller. $\pi_\text{loc}$ uses a student-teacher framework \cite{https://doi.org/10.48550/arxiv.2107.04034}, to adapt to different environment conditions as encoded in $z^\text{phys}$. In summary, the neural-network policy $\pi_\text{loc}$ returns the following action at each time $t$: $a_{t}^\text{loc} = \pi_\text{loc}(g_t, q_t, \dot{q}_t, z_t^\text{phys}, a_{t-1}^\text{loc}; c_t).$

\subsection{High-level Navigation Policy}
A navigation policy $\pi_\text{nav}$ is trained using PPO in cluttered corridors to provide a sequence of velocity commands $v^\text{des}= (v_x, v_y, v_\omega) \in \mathbb{R}^3$ to the low-level controller $\pi_\text{loc}$ so that the robot successfully navigates from its initial position to the goal region. The commanded gait $b$ is fixed to a trotting gait due to its simplicity and stability.
$\pi_\text{nav}$ takes as input a history $H$ of observations $o^\text{nav}$ and outputs a desired velocity command $v^\text{des}$. The observation contains the $\tt SE(2)$ pose of the robot $\rho$, the robot's velocity vector $(\dot{x}, \dot{y})$, and the previous action $v_\text{prev}^\text{des}$ output from the policy. At any time-step $t$, the high-level navigation policy outputs: $v_t^\text{des} = \pi_\text{nav}(\langle \rho_t, \dot{x}_t, \dot{y}_t, v^\text{des}_{t-1}\rangle_{t-H}^t).$
The history $H$ of observations (positions and velocities of the robot) enables the policy to interpret and remember interactions with the obstacles, which in turn aids in navigation. The reward function fuses (a) {\it goal reaching reward}: a sparse reward for the robot on reaching the goal region; (b) {\it distance penalty}: a penalty on the robot's distance from the goal region; (c) {\it time penalty}: incurred for every control step the robot does not reach the goal; (d) {\it wall collision penalty}: when the robot gets close to the walls; (e) {\it heading penalty}: if the robot's heading angle is beyond the threshold $\theta_\text{thresh}$ with respect to its initial heading.

\begin{figure*}[ht!]
    \centering
    \includegraphics[width=.99\textwidth]{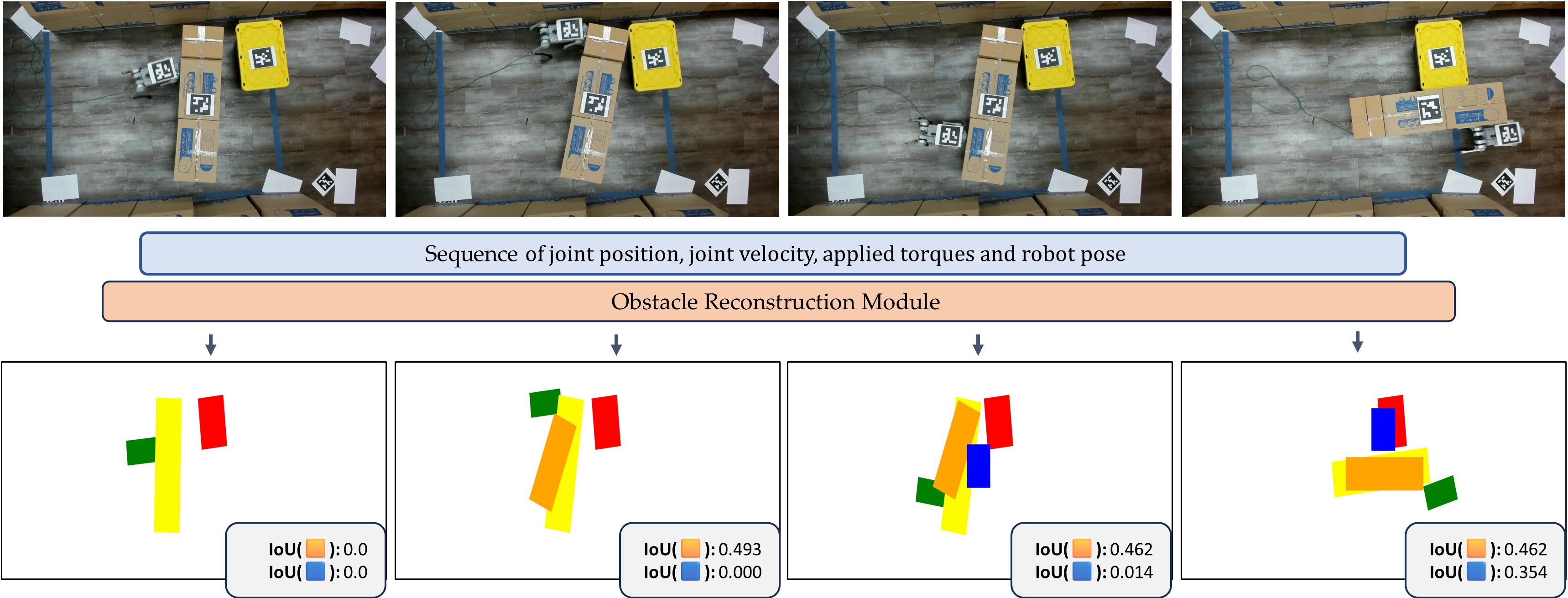}
    \caption{Given as input a sequence of the robot's joint positions, velocities, applied torques, and poses (Top), the Obstacle Reconstruction Module (ORM, middle) is a Transformer-based neural network that outputs the sizes and poses for the different movable (ground truth visualized in yellow) and static (ground truth visualized in red) obstacles in the scene (Bottom, best viewed in color). The reconstruction for the movable and static obstacles are visualized in orange and blue, respectively, and their corresponding Rotated Intersection Over Union (See: Section~\ref{sec:experiments}-C) values are reported. Higher values represent more accurate reconstructions.}
    \label{fig:obstacle-prediction}
    \vspace{-0.25in}
\end{figure*}


\subsection{Dataset Curation}
The Obstacle Reconstruction Module (ORM) is trained on a dataset collected by running the navigation policy $\pi_\text{nav}$ in a randomly generated simulation environment (see Sec.~\ref{sec:experiments} for details). 
The execution is terminated when the robot reaches the goal region, or after $T_\text{max}$ time-steps. The time-step when the goal is reached is denoted by $t_\text{goal}$. The robot's proprioceptive state $s$ and pose $\rho$, as well as the $N$ obstacle states $\mathcal{O}$, are recorded at every time step. A \textit{trajectory} $\gamma = \{s_t, \rho_t, \mathcal{O}^t\}_{t = 0}^{t=T}$ is defined as a sequence of observations, where $T=\text{min}(t_\text{goal}, T_\text{max})$. A dataset $\mathcal{D}$ is a collection of trajectories $\gamma_j$ under $M$ different navigation policies $\pi^i_\text{nav}$, $i \in \{1, 2, .., M\}$ in randomly generated cluttered environments.

Multiple sources of bias may arise in collecting $\mathcal{D}$:

\noindent\textbf{Navigation policy bias.} In an environment with multiple homotopic paths, $\pi_\text{nav}$ may be biased towards a single homotopy class (e.g., trajectories that favor the right-hand side of the environment). 
This may lead to recorded trajectories having limited robot-obstacle interactions.

\noindent \textbf{Early termination bias.} In different trajectories, the robot interacts with different numbers of obstacles in the scene. 
The trajectories collected from randomly generated environments are not uniformly divided with respect to the obstacle interaction modes, leading to an unbalanced dataset. 
For example, trajectories terminate faster in an environment with a single movable obstacle than in environments with multiple obstacles, increasing the number of trajectories recorded for the former.

\noindent \textbf{Contact mode bias.} Each trajectory can be categorized into different contact modes, e.g., no contact with any obstacle, direct contact with a static/movable obstacle, or direct contact with a movable obstacle along with indirect static obstacle contact. Imbalances between the modes can lead to bias.

To ensure a high-quality dataset, the curation procedure collects trajectories from $M$ different navigation policies to help cover multiple possible homotopic classes, increasing the diversity of interactions with the obstacles. Pruning trajectories based on contact mode frequency helps mitigate the early termination and contact mode biases. 

\subsection{Obstacle Reconstruction Module (ORM)}
ORM is a neural network consisting of a causal Transformer encoder \cite{devlin2019bert} followed by a fully-connected $\tt MLP$ decoder  that predicts a sequence of obstacle parameters for reconstruction from a sequence of proprioception inputs (Fig~\ref{fig:obstacle-prediction}). This network is trained on trajectories in $\mathcal{D}$. The input is a sequence of proprioception data $(s_t, \rho_t)_{t=0}^{T}$ -- joint positions, joint velocities, torques applied on the joints, and the robot pose in the world frame. 
For any trajectory in $\mathcal{D} $, let the time-steps of the first and last contact on obstacle $\mathcal{O}_i$ be denoted by $t_i^\text{first}$ and $t_i^\text{final}$, respectively. Obstacle $\mathcal{O}_i$'s contact window is defined as $\Delta_i^\text{con} = [t_i^\text{first}, t^\text{final}]$. $t^\text{final} = \text{max}_i(t_i^\text{final})$ denotes the time-step of the final contact with any obstacle in $\mathcal{O}$.
The obstacle parameters $\mathcal{O}_i$ (see Section III) are augmented to include the contact window information, $\Tilde{\mathcal{O}_i} = (\mathbb{I}^\text{contact}_i, \mathbb{I}^\text{static}_i, x_i, y_i, \theta_i, w_i, l_i)$. The binary variable $\mathbb{I}^\text{contact}_i$ indicates whether the obstacle is in its contact window (1), i.e. if the robot or another obstacle $\mathcal{O}_j$ is currently pushing against the obstacle $\mathcal{O}_i$ (where $i \neq j$), or not (0). This augmented $\mathcal{\Tilde{O}}^t$  is the ground-truth label for the learning process. The module outputs predictions $\{\mathcal{\hat{O}}^t\}_{t=0}^{t=T}$, where $\hat{\mathcal{O}}^t = \{\hat{\mathcal{O}}^{t}_{i}\}_{i=1}^{i=N}$,
$n$ is the number of obstacles in the scene and the prediction at time-step $t$ is a function of only the history of inputs up to time-step $t$ in the trajectory. The segments of the output sequence not in the contact window for each obstacle $\mathcal{O}_i$ are masked, so the network predicts $\mathcal{O}_i$'s parameters only during its contact window. Contact windows contain rich contact information for obstacle prediction.

The training loss for ORM is a combination of scaled binary cross-entropy $\texttt{BCE}$ losses and a mean squared error $\texttt{MSE}$ supervised learning objective. The loss function for a single sequence with $n \leq N$ interacted obstacles is:
 \begin{align*}
    \mathcal{L} = \sum_{i=1}^n \frac{1}{\vert \Delta_i^\text{con} \vert} \sum_{t=t_{i}^\text{first}}^{t^\text{last}} [
    \alpha_1~\texttt{BCE}(\mathbb{I}^\text{contact}_{t, i}, \hat{\mathbb{I}}^\text{contact}_{t, i}) \\
   +~\alpha_2~\texttt{BCE}(\mathbb{I}^\text{static}_{t, i}, \hat{\mathbb{I}}^\text{static}_{t, i}) \\
   +~\alpha_3~\texttt{MSE}((x_{t, i}, y_{t, i}, \theta_{t, i}), (\hat{x}_{t, i}, \hat{y}_{t, i}, \hat{\theta}_{t, i})) \\
   +~\alpha_4~\texttt{MSE}((w_{t, i}, l_{t, i}), (\hat{w}_{t, i}, \hat{l}_{t, i}))].
\end{align*}

\subsection{System Integration}
\noindent \textit{Offline Training} Locomotion policy $\pi_\text{loc}$ is first trained using PPO to follow velocity and gait commands. Once trained, $\pi_\text{loc}$ is used only in inference mode with a fixed gait for navigation. $M$ navigation policies $\pi_\text{nav}^i$, $i \in \{1, 2, ..., M\}$ are then trained using PPO in unknown cluttered environments using the low-level control policy $\pi_\text{loc}$. A balanced dataset $\mathcal{D}$ of trajectories is collected from $M$ different navigation policies in randomly generated cluttered environments. The ORM is then trained with a supervised objective to partially reconstruct the environments using only proprioception data.
\noindent \textit{Online Inference} For navigating through environments with unknown obstacles, $\pi_\text{nav}$ operates in inference mode, internally commanding $\pi_\text{loc}$ for robot locomotion. Each control step of the navigation process captures and records the robot's proprioception state in a sequence, preserving a complete history from the trajectory's start. This sequence feeds into the ORM, enabling partial, on-the-fly scene reconstruction. This proposed approach provides an novel method for real-time 2D-scene understanding during navigation.

\section{Experiments}
\label{sec:experiments}



\noindent \textbf{Environment Setup} For the evaluation, a 12 DoF Unitree Go1-EDU quadruped starts at one end of a walled $4$m$\times2$m corridor and must navigate to the goal region located at the other end of the corridor. 
\textit{Offline, during training}, obstacles $\mathcal{O}_i, i \in [1,N_\text{max}]$, are generated in IsaacGym by sampling their physical properties for each obstacle. The obstacles are then placed in random $\tt SE(2)$ configurations with their orientations fixed to $0$ with respect to the robot's initial orientation. This may or may not block the path(s) to the goal. When $N_\text{max} \ge 2$, the movable obstacle is placed in front of one or more static obstacles. No obstacles are spawned in the goal region. \textit{Domain randomization} of the obstacle's physical parameters aims to reduce the sim2real gap of both the navigation policy and the obstacle prediction module.
\begin{figure}[h!]
    \centering
    \vspace{-.1in}
    \begin{subfigure}{.325\columnwidth}
    \includegraphics[width=\textwidth]{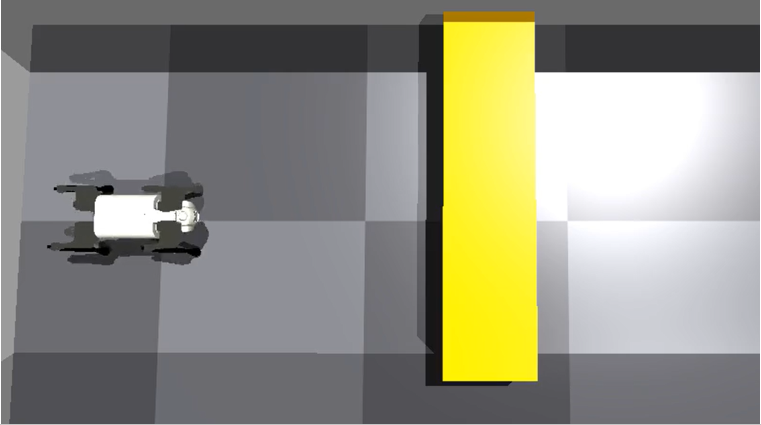}
    \end{subfigure}
    \begin{subfigure}{.325\columnwidth}
    \includegraphics[width=\textwidth]{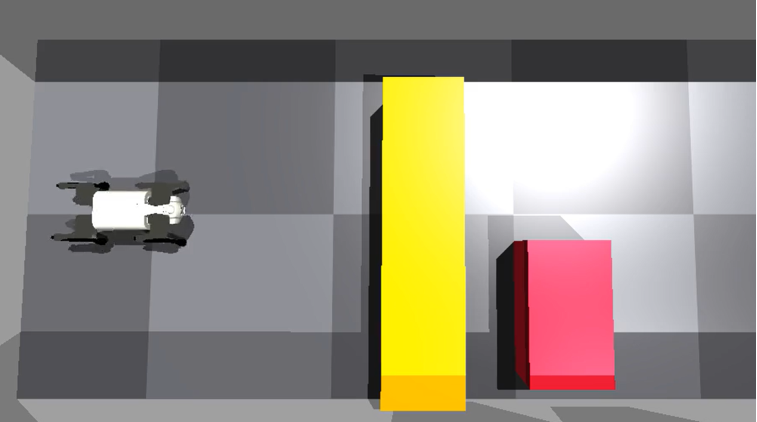}
    \end{subfigure} 
    \begin{subfigure}{.325\columnwidth}
    \includegraphics[width=\textwidth]{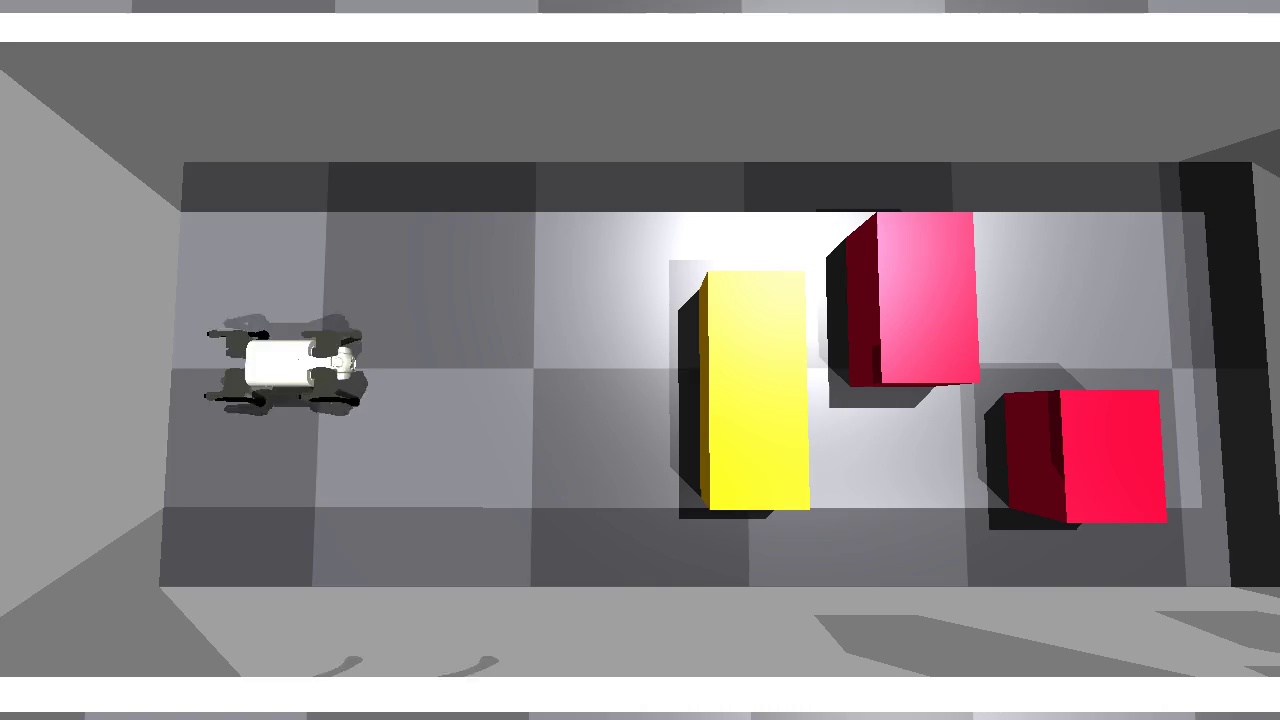}
    \end{subfigure} \\
    \begin{subfigure}{.325\columnwidth}
    \includegraphics[width=\textwidth]{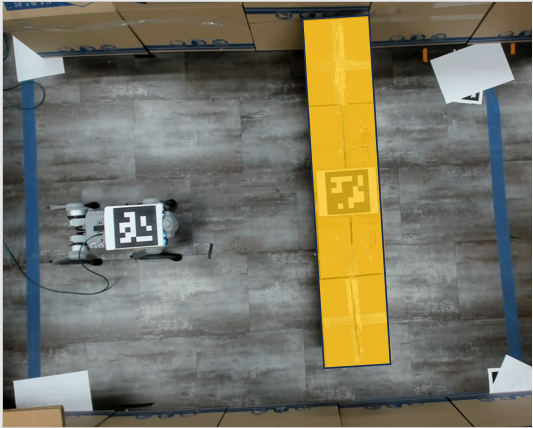}
    \end{subfigure}
    \begin{subfigure}{.325\columnwidth}
    \includegraphics[width=\textwidth]{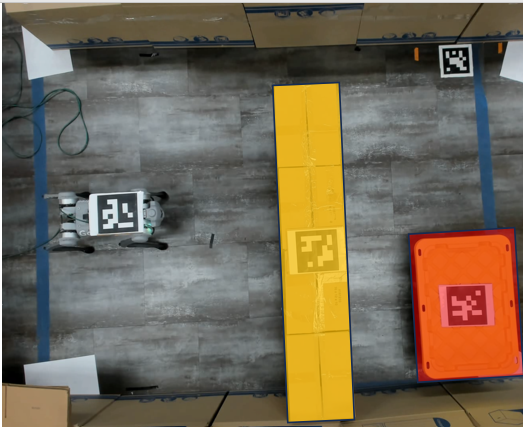}
    \end{subfigure}
    \begin{subfigure}{.325\columnwidth}
    \includegraphics[height=64pt, width=\textwidth]{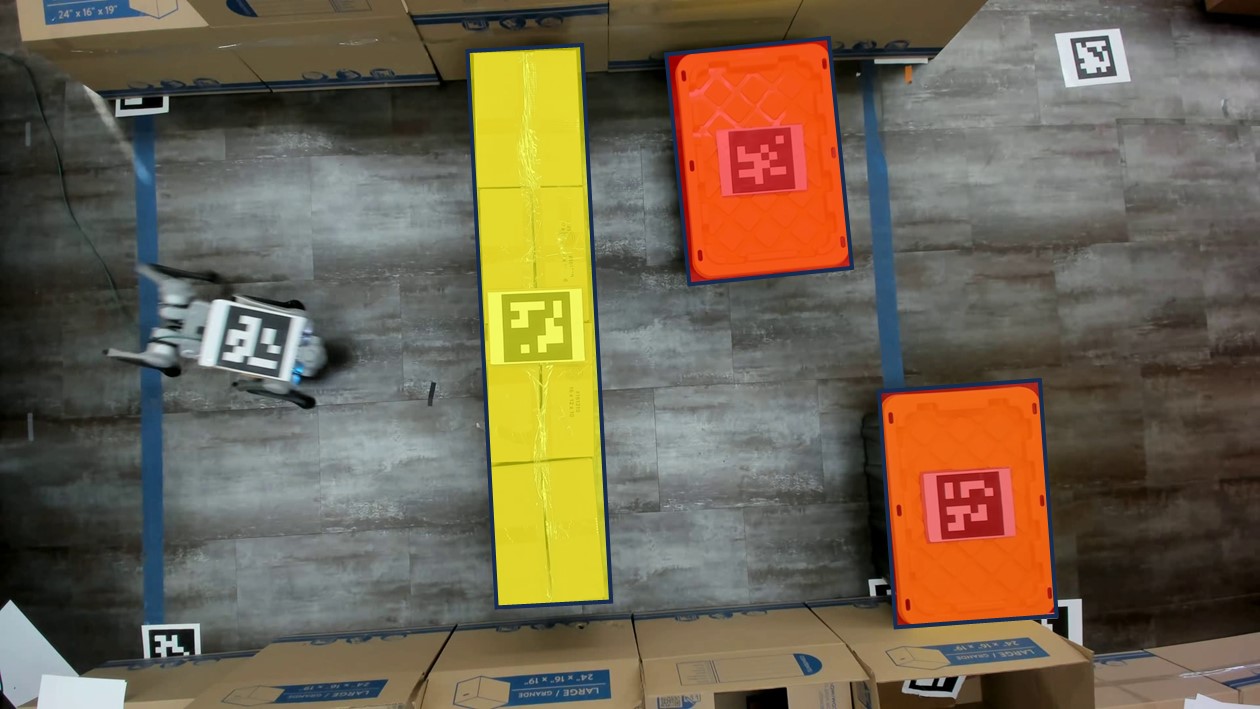}
    \end{subfigure} \\
    \vspace{-0.05in}
    \caption{Examples of Easy (Left), Medium (Middle) and Hard (Right) simulated (top) and real-world (bottom) scenarios considered in the evaluation.}
    \vspace{-.2in}
    \label{fig:environments}
\end{figure}

\textit{Online, during evaluations}, the environments are categorized into \textit{easy}, \textit{medium} and \textit{hard} with respect to the navigation task, where the number of obstacles $N$ is set to 1, 2, and 3 respectively. In the \textit{easy} environments, the robot may only directly interact with the obstacle, whereas, in the \textit{medium} and \textit{hard} ones, the robot may have both direct and indirect obstacle interactions.

\noindent \textbf{Training Details} The locomotion and navigation \textbf{policies}, $\pi_\text{loc}$ and $\pi_\text{nav}$, are trained on 4096 parallel IsaacGym environments for 30K and 1K iterations, respectively. $\pi_\text{loc}$ is trained to follow velocity commands in the range $[-0.4, 0.4]m/s$ at a step frequency of 50Hz. Outside this range, the robot is prone to falling when interacting with obstacles while executing $\pi_\text{nav}$. The input to $\pi_\text{nav}$ is the history of observations and actions, where the history is a moving window with a maximum length of $H = 30$s and the maximum episode length $T_\text{max} = 60$s. $\pi_\text{nav}$ provides commands to $\pi_\text{loc}$ at a step frequency of 25Hz. For better sim-to-real transfer of $\pi_\text{nav}$, the robot parameters are randomized during training. Multiple versions of $\pi_\text{nav}$ are trained on different seeds using different combinations of reward functions mentioned in Section IV-B.

A single {\bf Obstacle Reconstruction Module} (ORM) is trained on a dataset $\mathcal{D}$ of $180K$ trajectories curated from three different navigation policies $\pi_\text{nav}^i, i \in \{1,2,3\}$. The maximum number of obstacles spawned in the environment $N_\text{max} = 3$. The $\tt Transformer$ Encoder comprises four self-attention blocks with two attention heads each, followed by a two-layer $\tt MLP$ decoder. The inputs to the network are projected to learnable embedding of 512 dimensions with positional information. This network is trained for $20$ epochs on an NVIDIA RTX A4500 with 20GB of GPU memory. 

\noindent \textbf{Evaluation Metrics}
The following metrics measure the performance of the ORM.
They are calculated for each obstacle $\mathcal{O}_i$ independently at the final time-step $t_\text{final}$ in the contact window $\Delta^\text{con}_i$ (See Section IV-D).

\noindent {\bf Rotated Intersection over Union (IoU)} evaluates how well the predicted obstacle representation overlaps with the ground-truth.  For a single obstacle $\mathcal{O}_i$ at time-step $t$, it is defined as,
\begin{equation*}
    \text{IoU}(\mathcal{O}_{i}^t, \hat{\mathcal{O}}_{i}^t) = \frac{\text{Area}(\text{Geom}(\mathcal{O}_{i}^t)\text{ }\cap  \text{ } \text{Geom}(\hat{\mathcal{O}}_{i}^t) )}{\text{Area}(\text{Geom}(\mathcal{O}_{i}^t)\text{ }\cup  \text{ } \text{Geom}(\hat{\mathcal{O}}_{i}^t) )},
\end{equation*}
where Geom$(\mathcal{O}_i)$ is the geometry of the obstacle $\mathcal{O}_i$ in the workspace. 
The IoU for obstacle $\mathcal{O}_i$ at the final time-step $t_\text{final}$ in the contact window $\Delta^\text{con}_i$ is defined as $\text{IoU}_{i}^\text{final} = \text{IoU}(\mathcal{O}_{i}^{t_\text{final}}, \hat{\mathcal{O}}_{i}^{t_\text{final}})$.\\
{\bf Absolute Error} For each obstacle $\mathcal{O}_i$ at time-step $t$, (i) \textit{Obstacle pose error} is the the absolute error of the predicted $\tt SE(2)$ pose of the obstacle.  (ii) \textit{Obstacle shape error} is the absolute error for the dimensions of the obstacle. 
The absolute error at the final time-step $t_\text{final}$ in the contact time window $\Delta^\text{con}_i$ is defined as $\mathcal{E}_{i}^\text{final} = \vert \mathcal{O}_{i}^{t_\text{final}} -\hat{\mathcal{O}}_{i}^{t_\text{final}} \vert$.
It corresponds to the individual absolute errors in the position of the object in each axis, in its rotation, and length. 
\begin{figure*}[t!]
    \centering
    \includegraphics[width=\textwidth]{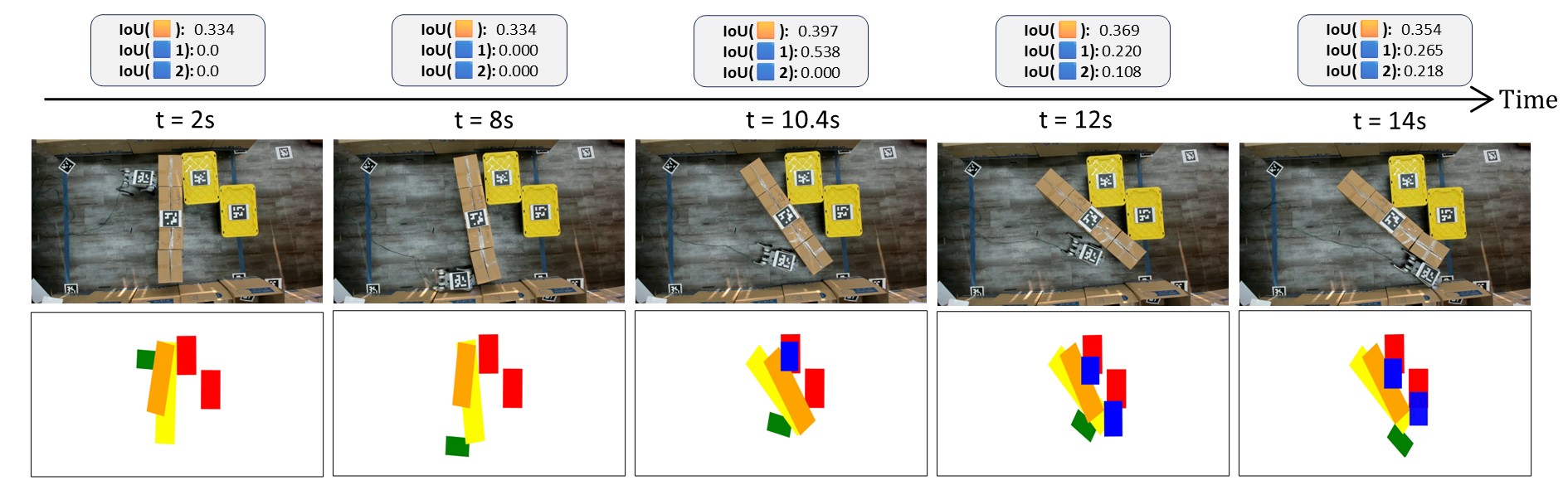}
    \caption{An example execution of \name with a real Unitree Go1, best viewed in color. (Top) Snapshots of the experiment at different timestamps. (Bottom) Obstacle reconstruction returned by \name. As the experiment progresses, the robot (green) comes into direct contact with the movable obstacle (ground truth pose in yellow) and indirect contact with the static obstacles (ground truth poses in red). The predictions during the contact window for the movable and static obstacles are visualized in orange and blue, respectively. }
    \label{fig:qual-real}
    \vspace{-0.25in}
\end{figure*}

\section{Results}
\label{sec:results}

\subsection{Simulation  Results}


\begin{table}[h!]
\centering
\def\arraystretch{1.2}%
\centering
\begin{tabular}{|c|c|c|c|c|c|c|}
\hline
\multirow{3}{*}{Category} & \multirow{3}{*}{Type} & \multicolumn{5}{c|}{Metrics} \\ \cline{3-7} 
                          &                               & \multirow{2}{*}{IOU $\uparrow$} & \multicolumn{3}{c|}{Pose Errors $\downarrow$} & \multirow{2}{*}{Shape $\downarrow$} \\ \cline{4-6}
                          &                               &     & x & y & $\theta$ &  \\ \hline
\multirow{2}{*}{Easy}     & Movable                       &  0.473   & 0.135  &  0.101  &   0.198       &  0.183     \\ \cline{2-7} 
                          & Static                        &  0.501   & 0.087  & 0.104  &  -        &   0.172    \\ \hline
\multirow{2}{*}{Medium}   & Movable                       &  0.496   &  0.115 & 0.095  &   0.201       &  0.162     \\ \cline{2-7} 
                          & Static                        &  0.331   & 0.430  & 0.169  & -      &    0.186    \\ \hline
\multirow{3}{*}{Hard}     & Movable                       &  0.481   &  0.128 &  0.108 &   0.214       &   0.172    \\ \cline{2-7} 
                          & Static 1                        & 0.432    &  0.091 & 0.138  &  -        & 0.117      \\ \cline{2-7} 
                          & Static 2                        & 0.404    &  0.094 &  0.151 &  -        & 0.120      \\ \hline
\end{tabular}
\caption{Quantitative results evaluating the proposed method in simulation. Performance outcomes are averaged across 1000 trials of \textit{Easy}, \textit{Medium}, and \textit{Hard} benchmarks that each contain $N_\text{max} = 1,2,3$ obstacles, respectively.}
\label{table:sim_results}
\vspace{-0.2in}
\end{table}

Table~\ref{table:sim_results} reports evaluations of the ORM using the aforementioned metrics on 1000 \textit{Easy}, 1000 \textit{Medium} and 1000 \textit{Hard} benchmarks. 
Across the benchmarks, the ORM returns a reasonably accurate estimation of the geometry of the obstacles evidenced by the average IoU at the end of the robot's contact as demonstrated qualitatively in Fig. \ref{fig:qual-real}.
In \textit{Medium} scenarios with a static obstacle, the robot may not be in direct or indirect contact with the static obstacle for a sustained time-period while completing the navigation task. Thus, the IoU reported in this case is relatively lower for the static obstacle. 
The reported reconstruction for the movable obstacle is surprisingly more accurate in the \textit{Medium} and \textit{Hard} scenarios.
The reason is that the robot is in direct contact with the obstacles for longer in the \textit{Medium} and \textit{Hard} benchmarks, which leads to more accurate predictions. The static obstacle's measured IoU is higher in the \textit{Hard} benchmark than the \textit{Medium} benchmark because there is a higher chance of direct interaction with one or the other static obstacle while navigating. The absolute errors further emphasize the trend observed in the IoU. Since the orientation of the static obstacles is fixed across the different benchmarks, their orientation error is omitted. 

\subsection{Real Robot Results}
Table~\ref{table:real_results} evaluates the ORM on independent trials of 20 \textit{Easy}, 20 \textit{Medium} and 5 \textit{Hard} benchmarks on the real-world setup. For each independent trial, a different navigation policy is uniformly sampled from $\{\pi_\text{nav}^i\}_{i=1}^{i=3}$. In the real-world setup, ArUco tags are used to estimate the ground-truth poses of the objects for computing the evaluation metrics. The robot's pose is used as input to the ORM, and can be replaced with onboard localization, e.g., by using an IMU.

\begin{table}[h!]
\vspace{-.05in}
\centering
\def\arraystretch{1.2}%
\centering
\begin{tabular}{|c|c|c|c|c|c|c|}
\hline
\multirow{3}{*}{Categ.} & \multirow{3}{*}{Type} & \multicolumn{5}{c|}{Metrics} \\ \cline{3-7}
                          &                               & \multirow{2}{*}{IOU $\uparrow$} & \multicolumn{3}{c|}{Pose Errors $\downarrow$} & \multirow{2}{*}{Shape $\downarrow$} \\ \cline{4-6}
                          &                               &     & x & y & $\theta$ &  \\ \hline
\multirow{2}{*}{Easy}     & Movable                       &  0.271   & 0.495  & 0.151  &   0.333      &  0.341     \\ \cline{2-7} 
                          & Static                        &  0.449   &  0.095 & 0.162  &   -       &  0.099     \\ \hline
\multirow{2}{*}{Medium}   & Movable                       &   0.277  &  0.340 &  0.195  & 0.407         &   0.303    \\ \cline{2-7} 
                          & Static                        &   0.212  & 0.185 & 0.230  &  -   &   0.225    \\ \hline
\multirow{3}{*}{Hard}     & Movable                       &   0.371  &  0.203 & 0.106  &  0.182        &  0.282     \\ \cline{2-7} 
                          & Static 1                        &  0.210 & 0.295  & 0.408  &  -        &  0.192     \\ \cline{2-7} 
                          & Static 2                       &   0.198 &  0.220 &  0. 494 & -        &  0.157     \\ \hline
\end{tabular}
\caption{Quantitative evaluation of \name on the real robot. Performance outcomes are averaged across 10 independent trials with a movable obstacle and 10 trials with a static obstacle in the \textit{Easy} category, followed by 20 trials with both obstacles in the \textit{Medium} category, and 5 trials with three obstacles in the \textit{Hard Category}.}
\label{table:real_results}
\vspace{-0.1in}
\end{table}

The results from the real-world trials confirm the trends observed in the simulation trials. However, multiple factors contribute to a certain decline in the measured metrics -- the physical properties of the obstacles, e.g., mass, friction, and restitution, are assumed to be uniformly distributed across the obstacle geometry in simulation, but this is not the case in the real-world setup. The real-world obstacles are also not perfectly box-shaped, contributing to a distribution mismatch between the obstacles and the robot compared to those in the simulation. The \textit{Hard} benchmark contains trajectories that have very minimal direct interaction with the static obstacles in the environment and this reflects in the measured IoU. This is comparable to the evaluated IoU for the \textit{Medium} benchmark. 

Fig~\ref{fig:qual-real} showcases an example execution of \name from a real robot trial on a \textit{Hard} benchmark. The robot makes first contact with the movable obstacle at $t = 2s$ and returns an initial estimate of its pose and shape. At $t = 8s$, $\pi_\text{nav}$ directs the robot to the other end of the movable box, creating an opening. As the robot continues pushing the movable box, it enables the ORM to hypothesize about a possible static obstacle at $t = 10.4s$. When the movable box can no longer be pushed at around $t = 12s$, the ORM reconstructs the second static obstacle nested behind the first one. However, since it does not have enough information about how it has manipulated the movable obstacle, the corresponding IoU decreases. The final reconstructions are obtained at the end of the contact window ($t = 14s$).
\vspace{-0.2in}

\begin{figure}[h!]
    \centering
    \includegraphics[width=\columnwidth]{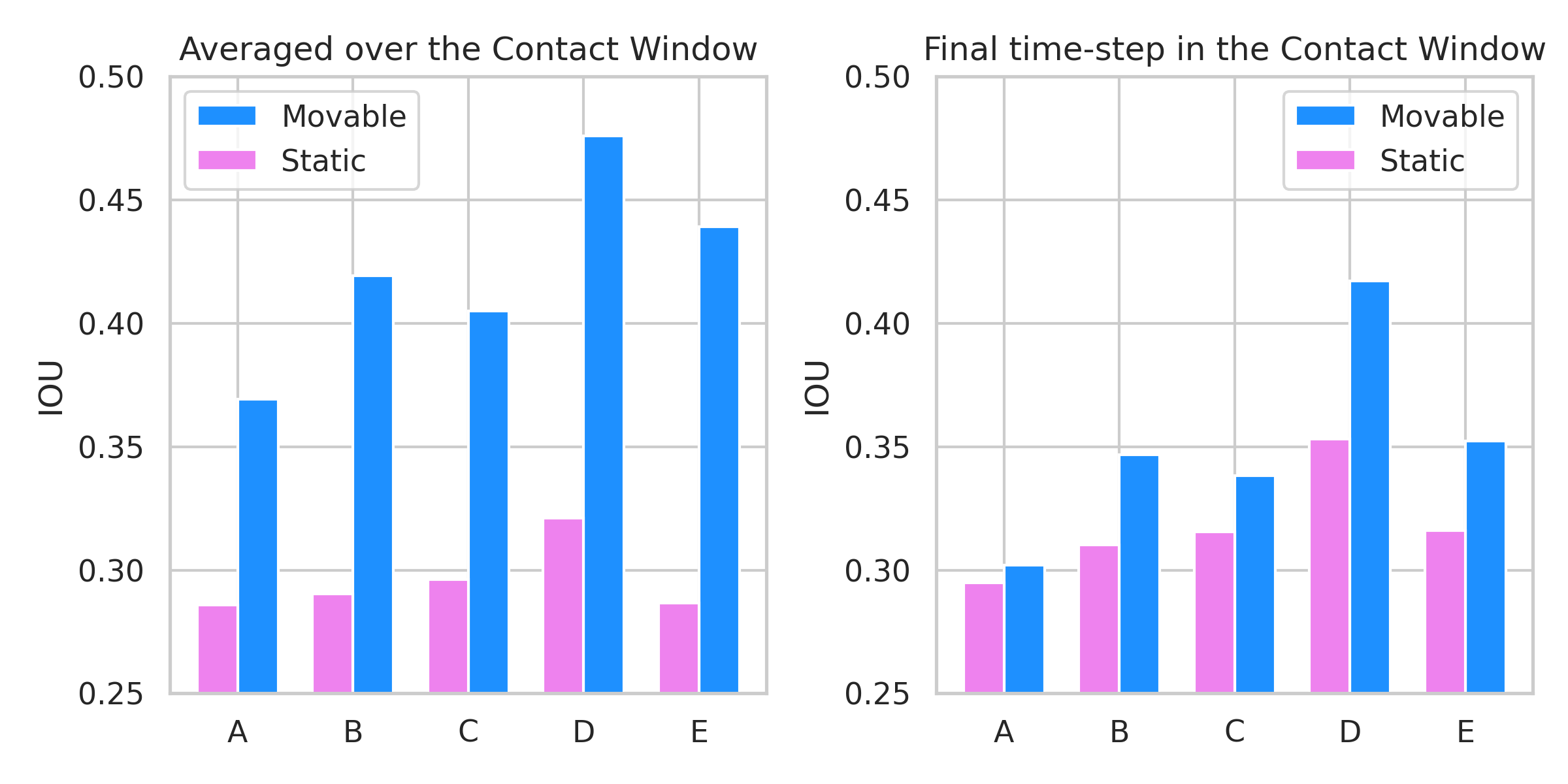}
    \caption{A: $\{q\}$,  B: $\{q, \dot{q}\}$, C: $\{q, \dot{q}, \tau\}$, \textbf{D:} $\{q, \dot{q}, \tau, \rho\}$, E: $\{\tau, \rho\}$. The plots show an ablation study on the ORM. The ORM trained with different inputs affects the reconstruction performance measured with the IoU metric.}
    \label{fig:input_ablation}
    \vspace{-0.2in}
\end{figure}
\noindent \textbf{Ablation Studies.} Fig~\ref{fig:input_ablation} evaluates the importance of the different inputs provided to the ORM using the IoU metric on the same evaluation dataset as Table~\ref{table:sim_results}. For the ablation study the ORM is trained on environments with $N_\text{max} = 2$. The networks $A, B, C, D$ receive as input the history of joint positions $q$, joint velocities $\dot{q}$, commanded joint torques $\tau$, and the robot pose $\rho$ incrementally. Network $E$ receives as input only the commanded joint torques and the robot's pose. For the movable obstacles, whose features are indicated by the changes in the robot's pose relative to the commanded joint torque, networks $D$ and $E$ perform best in accurately reconstructing them. The IoU of the static obstacles displays a marked improvement when the joint torques are introduced as input. Removing the joint position and velocity information significantly declines reconstruction performance, emphasizing their importance.
\section{Conclusion}
\label{sec:conclusion}
Robots that navigate unknown environments benefit from reconstructing their surroundings for planning their actions accordingly. While standard reconstruction methods rely on vision, \name is a novel method for 2D scene reconstruction for legged robots which uses only a history of proprioceptive data. This sensing modality is useful in places where vision is not helpful, such as those encountered in search and rescue missions or when obstacles are partially or fully occluded. \name\ employs a transformer-based obstacle reconstruction module, trained in simulation, to map the history of proprioceptive data into an encoded representation of the obstacle geometries in the environment. Experiments are carried out to first evaluate \name\ in simulation and then with a real Unitree Go1 robotic dog navigating an environment with static and movable objects. The evaluation indicates that \name\ can not only detect object locations and sizes but also whether they are movable or static. It can also detect the properties of fully occluded objects by interacting with the occluding frontal objects.

\name, however, has some limitations that are worth investigating in future works. Firstly, all experiments were conducted in environments that contained only box-shaped objects. It is interesting to extend and evaluate \name\ on other forms of objects. Furthermore, \name\ was not tested on identifying other physical properties of objects beyond their geometries and whether they are movable. Finally, combining \name\ with reconstruction methods that rely on other modalities, such as touch sensing, can improve the ability to reconstruct scenes.
\bibliographystyle{format/IEEEtran}
\bibliography{refs.bib}
\end{document}